\documentclass[sigconf]{acmart}
\usepackage{xcolor}
\usepackage[utf8]{inputenc}
\DeclareUnicodeCharacter{2212}{-}
\AtBeginDocument{%
  }

\setcopyright{acmlicensed}
\copyrightyear{2026}
\acmYear{2026}
\acmDOI{XXXXXXX.XXXXXXX}
\acmConference[KDD '27]{The 33rd ACM SIGKDD Conference on Knowledge Discovery and Data Mining}{August, 2027}{San Jose, CA, USA}

\acmISBN{978-1-4503-XXXX-X/2027/08}

\begin{document}

\title[Benchmarking the State of AI for Crop Yield Forecasting: A Global Assessment Across Modeling Paradigms]{Benchmarking the State of AI for Crop Yield Forecasting: \newline A Global Assessment Across Modeling Paradigms}

\author{Michiel Kallenberg}
\orcid{0000-0002-4661-6674}
\correspondingauthor
\affiliation{%
  \institution{Artificial Intelligence Group, Wageningen University and Research}
  \city{Wageningen}
  \country{the Netherlands}
}
\email{michiel.kallenberg@wur.nl}

\author{Philip Janz}
\affiliation{%
  \institution{Earth Observation for Agroecosystems, Agroscope}
  \city{Zurich}
  \country{Switzerland}
}
\affiliation{%
  \institution{Potsdam Institute for Climate Impact Research (PIK), Germany}
  \city{}
  \country{}
}
\email{philip.janz@agroscope.admin.ch}

\author{Christoph Jörges}
\affiliation{%
  \institution{Department of Geography, Ludwig-Maximilians-Universität Munich}
  \city{Munich}
  \country{Germany}
}
\email{c.joerges@lmu.de}

\author{Vageesh Saxena}
\affiliation{%
  \institution{Department of Advanced Computing Sciences, Maastricht University}
  \city{Maastricht}
 \country{the Netherlands}
 }
\email{v.saxena@maastrichtuniversity.nl}

\author{Pratishtha Poudel}
\affiliation{%
  \institution{Department of Agronomy, Purdue University}
  \city{West Lafayette}
  \state{Indiana}
  \country{USA}
}
\email{ppoudel@purdue.edu}

\author{Mohammed Musthafa Rafi}
\affiliation{%
  \institution{Department of Computer Science, Iowa State University}
  \city{Ames}
  \state{Iowa}
  \country{USA}
}
\email{mohd7@iastate.edu}

\author{Ioannis N. Athanasiadis}
\affiliation{%
  \institution{Artificial Intelligence Group, Wageningen University and Research}
  \city{Wageningen}
  \country{the Netherlands}
}
\email{ioannis.athanasiadis@wur.nl}

\renewcommand{\shortauthors}{Kallenberg et al.}

\begin{abstract}
Accurate crop yield forecasting is essential for food security, agricultural policy, and commodity markets. Although statistical, process-based, and machine learning approaches have all been proposed, their relative strengths remain poorly understood due to the lack of large-scale, standardized comparisons. We address this gap through a comprehensive benchmark of five modeling paradigms—statistical baselines, process-based models, conventional feature-engineered machine learning, deep sequence models, and tabular foundation models—using the CY-Bench benchmark spanning 63 country–crop datasets (>12,000 administrative regions) across six continents. Beyond conventional pooled accuracy metrics, we introduce a multi-dimensional evaluation framework that separately quantifies overall, spatial, temporal, and anomaly prediction skill. Across multiple crops and countries, feature-engineered machine learning and tabular foundation models achieve the strongest overall predictive performance. Data-driven models reduce NRMSE in 72\% of countries growing maize and 78\% of countries growing wheat, with median reductions of 8.8\% and 11.9\%, respectively. Across all modeling paradigms, spatial prediction is substantially easier than temporal and anomaly prediction. While data-driven models markedly improve temporal and anomaly prediction over statistical baselines, accurately forecasting year-to-year yield variability remains challenging. These findings suggest that future progress in large-scale crop yield forecasting will depend on the improved characterization of the environmental and management factors driving inter-annual yield variability, founded on effective representations for crop yield forecasting systems that support agricultural decision-making, early-warning applications, and food security planning.
\end{abstract}


\maketitle

\section{Introduction}
Crop yield forecasts are crucial in agricultural science. For example, when drought threatens maize production in East Africa or heat waves reduce wheat yields in South Asia, the speed and accuracy of yield predictions influence whether humanitarian organizations pre-position food aid, governments initiate emergency imports, and commodity markets accurately price risk \citep{miron2023influence}. However, despite decades of research, crop yield forecasts remain subject to substantial uncertainty arising from systems complexity, data limitations and modeling approaches, limiting their usefulness for agricultural decision-making, especially under a changing climate \citep{vandervelde_nisini2019}. The crop yield forecasting community lacks a systematic understanding of which modeling paradigms perform most reliably across diverse geographic regions, environmental conditions, crop types, and stages of the growing season \citep{sweet2025transdisciplinary, kallenberg2026cybench}. Furthermore, a wide range of modeling approaches exists for crop yield forecasting, including statistical models, process-based crop models, machine learning, deep learning, and foundation models, as well as hybrid approaches \citep{basso_liu2019, schauberger_etal2020, gavassorita_etal2023, han2025}. While several studies report improvements, few works offer results across several crops and countries~\citep{paudel_etal2021, Paudel2022,Paudel2023}. Typically, comparisons are restricted to a single country, crop, or modeling paradigm, limiting conclusions about generalizability across diverse agricultural systems \citep{sweet2025transdisciplinary, kallenberg2026cybench, miranda2026yieldsat, davis2025harveststat}.

Moreover, crop yield prediction poses a broader challenge for machine learning, as models must generalize across both spatial and temporal dimensions. Yield variability is largely determined by persistent spatial differences in productivity between administrative regions, whereas year-to-year variability within regions is smaller and primarily influenced by weather anomalies and extreme events \citep{frieler2017understanding}. Consequently, while a model may demonstrate strong overall predictive performance by replicating long-term spatial yield patterns, it may still fail to capture the inter-annual variability. For example, a model may correctly predict that one region consistently out-yields another because of more favorable soils or management, yet may still fail to predict severe yield losses during a drought year. While such a model achieves high overall accuracy, it provides limited value for early-warning applications that depend on anticipating inter-annual anomalies, demonstrating the importance of sufficient spatial and temporal evaluation schemes.

To address these needs, this research presents a comprehensive benchmark of modeling paradigms for sub-national crop yield forecasting based on the recently introduced CY-Bench dataset \citep{kallenberg2026cybench}. CY-Bench combines standardized environmental predictors with harmonized sub-national yield observations spanning multiple crops, countries, and agro-ecological conditions, providing a common basis for rigorous cross-paradigm evaluation.

The main contributions of this research are threefold. First, we conduct a comprehensive benchmark that compares statistical baselines, process-based models, feature-engineered machine learning, deep sequence models, and tabular foundation models within a unified experimental framework for sub-national crop yield forecasting at near-global coverage. Second, we introduce a multi-dimensional evaluation framework that separately quantifies overall, spatial, temporal, and anomaly prediction skill, revealing complementary strengths that remain hidden when evaluation relies solely on pooled accuracy metrics. Third, we demonstrate that current forecasting performance is primarily limited by the ability to predict inter-annual variability rather than persistent spatial productivity patterns, suggesting that future progress will depend increasingly on learning effective representations of the underlying environmental and management drivers.

Together, these contributions provide a strong empirical foundation for developing crop yield forecasting systems that support agricultural decision-making, early-warning applications, and food security planning.

\section{Related Work}
Crop yield forecasting has traditionally relied on statistical regression techniques and process-based crop models that simulate crop growth using biophysical principles and agronomic knowledge \citep{basso_liu2019, leng2020predicting, schauberger_etal2020}. More recently, machine learning methods, including tree-based ensembles \citep{rafi2026optimized}, recurrent neural networks \citep{fan2022gnn}, convolutional neural networks \citep{nevavuori2019crop}, transformer architectures \citep{hasan2026vita}, and foundation models \citep{sabo2026rows}, have demonstrated increasingly strong predictive performance. Most published studies, however, evaluate only specific families of models \citep{patil2025comparative, rafi2026benchmarking} and typically focus on a single country \citep{leng2020predicting, srivastava2022winter, corcoran2026comprehensive}, crop \citep{ashfaq2025predicting, miranda2026yieldsat}, or modeling paradigm \citep{muller2019global, ma2026harvesting}. Consequently, reported improvements are often difficult to generalize across agricultural systems and environmental conditions.

Systematic model intercomparison has long been recognized as an important tool for understanding model strengths, weaknesses, and uncertainty. The Agricultural Model Intercomparison and Improvement Project (AgMIP) and its Global Gridded Crop Model Intercomparison (GGCMI) \citep{muller2019global} established standardized protocols to compare process-based crop models under common climate forcing, yielding important insights into structural uncertainty and climate impacts on agricultural production. However, these efforts are limited to process-based crop models and yet do not address comparisons with modern data-driven approaches.

Recent benchmark datasets have established standardized resources for machine-learning-based crop yield forecasting. While some benchmarks emphasize field-scale prediction using high-resolution satellite imagery and multimodal observations \citep{lin2024open, miranda2026yieldsat}, others, such as CY-Bench, focus on regional crop yield forecasting across multiple crops and countries. This makes CY-Bench particularly well suited for controlled comparisons of diverse modeling paradigms under identical data conditions. Existing benchmark studies primarily compare machine-learning models with process-based approaches \citep{leng2020predicting, liu-et-al2022}, but their scope is typically restricted to specific model families, a single country, or a single crop. To our knowledge, no previous work has systematically compared statistical baselines, process-based models, conventional machine learning, deep sequence models, and tabular foundation models within a unified experimental framework while simultaneously evaluating overall, spatial, temporal, and anomaly prediction skill.

\section{Dataset}
We conduct experiments on the CY-Bench dataset \citep{kallenberg2026cybench}, a global benchmark that integrates publicly available yield statistics with environmental predictors relevant to crop growth, including weather variables, soil moisture indicators, remote-sensing vegetation indices, and soil properties, for sub-national crop yield forecasting. The dataset covers maize in 36 countries and wheat in 27 countries across six continents, providing yield observations for more than 12,000 sub-national administrative regions. Predictor variables are spatially intersected with crop type maps, aggregated to regional units, and temporally aligned with crop-growing seasons, enabling consistent evaluation across diverse agro-ecological conditions.

\begin{figure*}[!h]
  \centering
 \includegraphics[width=\textwidth]{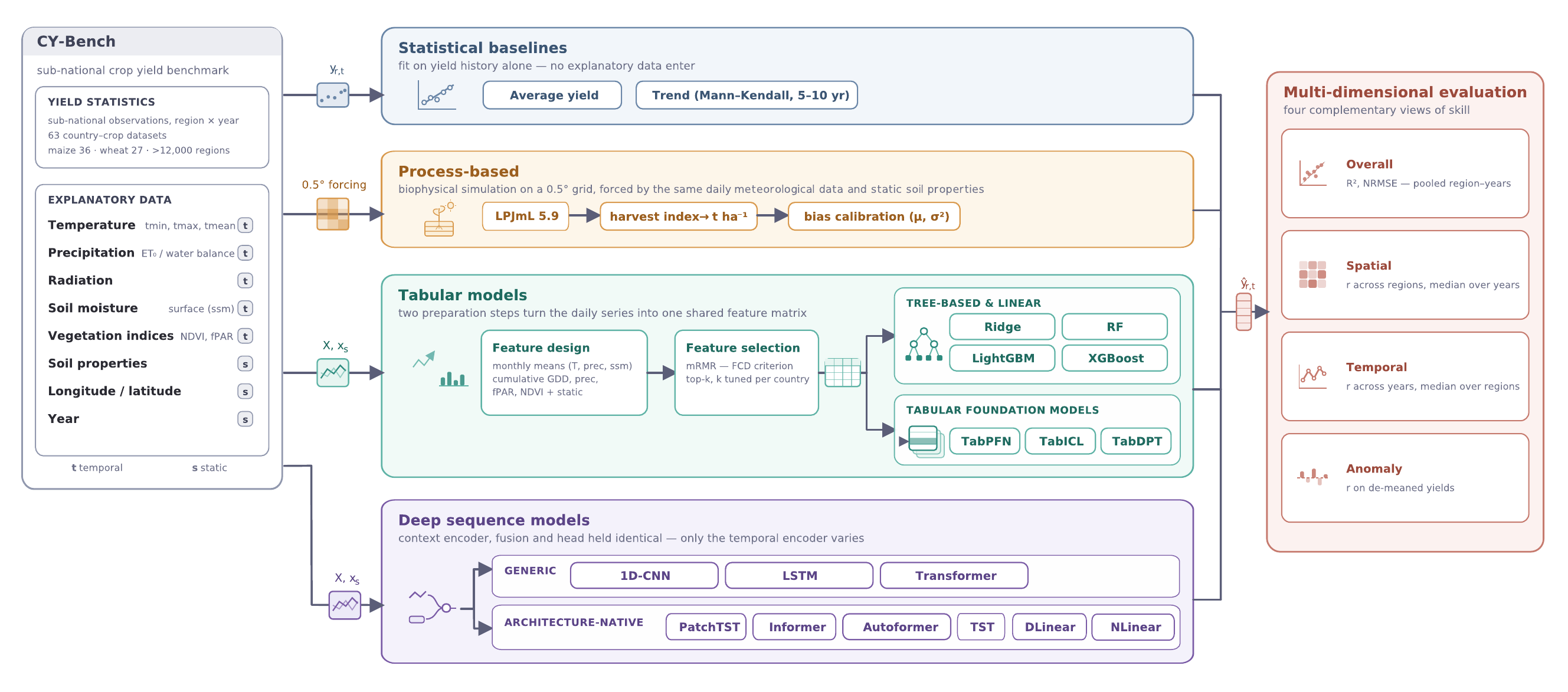}
  \caption{Overview of the benchmark. CY-Bench provides harmonized sub-national crop yield observations and environmental predictors for benchmarking statistical, process-based, tabular, and deep sequence models under a common experimental protocol. Performance is evaluated using four complementary dimensions: overall, spatial, temporal, and anomaly prediction.}
  \Description{Overview of workflow.}
  \label{fig:overview}
\end{figure*}

To improve robustness, we extend the CY-Bench yield target with a standardized quality-control step. For each country–crop dataset, region–year observations are compared with a locally fitted quadratic yield trend. Observations are flagged as outliers when they exceed the fitted trend by more than three standard deviations, and are removed from both training and evaluation. Consistent with recent work on harmonizing sub-national crop statistics \citep{lee2025harveststat}, we only filter out positive outliers, as unusually low yields may correspond to genuine crop failures and are therefore retained.

\section{Model paradigms}

Figure~\ref{fig:overview} summarizes the benchmark workflow. We evaluate models from five modeling paradigms: statistical baselines, process-based models, feature-engineered machine learning, (deep) sequence models, and (tabular) foundation models.

\subsection{Statistical baselines} 
We include two statistical baselines commonly used in crop yield forecasting:

\emph{Average yield} represents the historical mean yield of an administrative region computed over the training years. This baseline captures persistent regional productivity differences, but ignores temporal variability arising from weather, management, climate trends, or other environmental factors.

\emph{Trend yield} extends the regional-mean baseline by allowing for gradual changes in regional productivity over time. For each administrative region and forecast year, the model considers yield histories from the training period and searches over window lengths of 5--10 years.
For each candidate window, a Mann--Kendall test is applied to assess whether yields exhibit a significant monotonic trend; the window with the strongest evidence of trend is selected. A linear regression of yield on calendar year is then fitted to this window and extrapolated to the target year, capturing long-term technological change, management intensification, and slow climatic shifts. If no significant monotonic trend is detected, the prediction falls back to the regional mean.

\subsection{Process-based modeling}
Process-based crop models simulate plant growth using biophysical representations of crop development, carbon assimilation, and resource limitations \citep{noia2025crop}. These models are typically driven by meteorological forcing data and simulate crop growth throughout the entire growing season, producing yield estimates at the end of the season once physiological maturity is reached. 

As a process-based baseline, we employ the dynamic global vegetation model LPJmL (Lund–Potsdam–Jena managed Land) \citep{schaphoff2018lpjml4}, an agro-ecosystem model that simulates crop growth, natural vegetation dynamics, carbon and water fluxes, and the terrestrial nitrogen cycle to estimate crop yields for different crop functional types (CFT) \citep{von2018implementing, schaphoff2018lpjml4}. Crop growth is simulated from daily meteorological forcing variables, including temperature, precipitation, radiation, and yearly atmospheric CO$_2$ concentrations. Photosynthesis, respiration, biomass allocation, and phenological development are simulated on a daily time step, with phenology governed by temperature-driven growing degree day (GDD) accumulation. Net primary production is partitioned among plant organs according to CFT-specific allocation rules.

We use LPJmL version 5.9 on a global 0.5° grid. To ensure a fair comparison with the machine learning models, we feed LPJmL with the gridded meteorological inputs of the CY-Bench dataset, derived from the AgERA5 reanalysis \citep{boogaard_etal2022}, including daily temperature, precipitation, solar radiation, and wind speed. Soil properties, including soil type and pH, are obtained from the Harmonized World Soil Database \citep{nachtergaele2010harmonized}, while crop distributions are prescribed using the WorldCereal crop masks \citep{vantricht_etal2023} included in the CY-Bench raw data. Although LPJmL distinguishes between irrigated and rainfed production systems, all crop masks are interpreted as rainfed to match the assumptions adopted throughout this benchmark. We retain the default LPJmL management configuration to provide a globally consistent, agronomically realistic, process-based baseline. 

Finally, LPJmL outputs harvested vegetation carbon, which we convert to crop yield using the standard harvest-index formulation $Y_{\mathrm{LPJmL}} = C_h/(c\times dm)\times10^{-2}$, where $c=0.45$ is the biomass carbon content and $dm$ is the CFT-specific dry matter content (0.85 for maize and 0.87 for wheat) \citep{hermelink2021}. To facilitate a fair comparison with the other forecasting models, we further calibrate the simulated yields by matching the observed regional mean and variance computed from the training years, thereby correcting for systematic biases while preserving interannual variability.

LPJmL has been extensively evaluated within GGCMI \citep{mueller2017ggcmeval}, making it a well-established process-based baseline. However, its relatively coarse spatial resolution limits its ability to represent sub-grid heterogeneity and capture local yield variability.

\subsection{Data-driven models}

This research evaluates several data-driven models for crop yield forecasting formulated as a point-yield prediction task. Given a set of predictors, including weather variables, soil moisture indicators, remote-sensing vegetation indices, and soil properties, each model learns to predict yearly crop yield of a specific region within a country as a regression problem.
The evaluated models differ primarily in their use of temporal information and domain knowledge. Tabular models operate on engineered agronomic features and comprise conventional machine learning and tabular foundation models, whereas sequence models learn representations directly from raw temporal predictors without predefined feature aggregation.

\subsubsection{Feature-Engineered Machine Learning Models}
Feature-based models incorporate domain knowledge by aggregating the raw time-series predictors into agronomically meaningful summary statistics. These include monthly averages of minimum temperature (\textit{tmin}), maximum temperature (\textit{tmax}), mean temperature (\textit{tmean}), precipitation (\textit{prec}), climatic water balance (\textit{prec} $-$ \textit{ET\textsubscript{0}}), and surface soil moisture (\textit{ssm}). End-of-month values are computed for cumulative \textit{G}rowing \textit{D}egree \textit{D}ays (\textit{GDD}), cumulative precipitation, cumulative \textit{f}raction of absorbed \textit{P}hotosynthetically \textit{A}ctive \textit{R}adiation (fPAR), and cumulative  \textit{N}ormalized \textit{D}ifference \textit{V}egetation \textit{I}ndex (NDVI). Static soil and location covariates are encoded separately. To reduce redundancy and retain the most informative predictors, we perform feature selection using the minimum Redundancy Maximum Relevance (mRMR) \citep{peng2005feature} criterion during the screening phase. The resulting feature subsets are then used to train Ridge Regression, Random Forest (RF), LightGBM, and XGBoost models.

\subsubsection{Foundation Models}
Like the conventional machine learning models, tabular foundation models operate on the same engineered feature representation. Unlike conventional machine learning, however, they leverage representations learned during large-scale pretraining on external tabular datasets rather than learning solely from the CY-Bench training data. This pretraining enables them to learn general representations that transfer across downstream prediction tasks, often requiring substantially less domain-specific training data. To ensure a fair comparison, all foundation models evaluated in this study are restricted to the predictors available in CY-Bench and do not use any additional data sources during downstream training or inference.

We evaluate three tabular foundation models: TabICL \citep{qu2025tabicl}, TabPFN \citep{hollmann2022tabpfn}, and TabDPT \citep{ma2026tabdpt}, which adapt their pretrained representations to crop yield prediction via in-context learning \citep{dong2024survey}.

\subsubsection{Deep Sequence Models}
Unlike feature-engineered models, which rely on manually derived agronomic summaries, sequence models learn representations directly from daily weather and remote-sensing observations aligned with the crop calendar. To separate the contribution of temporal representations in different encoders, we distinguish between \emph{generic late-fusion encoders} and \emph{architecture-native encoders}. That said, all our neural models follow a common late-fusion architecture consisting of a static context encoder, a temporal encoder, a fusion module, and a regression head. By keeping the static context encoder, fusion module, and regression head identical across models, differences in predictive performance can be attributed primarily to the temporal encoder.

\noindent \paragraph{1. Generic late-fusion encoders:}
These models provide a controlled comparison of temporal encoding strategies within a common late-fusion architecture. Each variant first tokenizes the daily multivariate sequence using a learnable 1D convolution (kernel size~$p$, stride~1) followed by max-pooling (window and stride~$p$), producing a sequence of tokens with embedding dimension~$D$. During hyperparameter optimization, we consider $p \in {1,3,5,7,9}$, where $p{=}7$ corresponds to weekly tokens. We evaluate three variants that differ only in the temporal encoder and pooling rule: LSTM (stacked LSTM, last-token pooling), CNN (temporal Conv1D layers, attention pooling), and Transformer (standard Transformer encoder, attention pooling, with learned sin/cos day-of-year positional features). The temporal embedding $\mathbf{h}^{\mathrm{tmp}}_i \in \mathbb{R}^{D}$ is fused with the static context representation before yield prediction.

\noindent \paragraph{2. Architecture-native encoders:}
We additionally evaluate several state-of-the-art time-series forecasting architectures by integrating their temporal backbones into our late-fusion framework: PatchTST \citep{nie2022time}, Informer \citep{zhou2021informer}, Autoformer \citep{wu2021autoformer}, the Time-Series Transformer (TST) \citep{alexandrov2020gluonts}, and the linear decomposition models DLinear and NLinear \citep{zeng2022transformerseffectivetimeseries}. PatchTST, Informer, Autoformer, and TST are implemented using the Hugging Face \texttt{transformers} library \citep{wolf2020huggingfacestransformersstateoftheartnatural} and retain their native tokenization, positional encoding, and attention mechanisms. DLinear and NLinear preserve their original channel-wise linear formulations. For all architecture-native encoders, we apply mean pooling over the temporal dimension to obtain a fixed-length representation, which is subsequently projected to dimension~$D$ before fusion with the static context representation. Although these architectures were originally developed for multi-step time-series forecasting, we repurpose them here as temporal representation learners for scalar crop yields.


\section{Experimental setup and evaluation}
\label{sec:protocol}
This research implements a two-phase evaluation protocol comprising a Screening Phase and a Walk-Forward Phase. The Screening Phase applies a fixed temporal split: for each country-crop dataset, the last five available years are allocated to the test set, the two preceding years to the validation set, and all earlier years to the training set. As temporal coverage varies across countries, the specific calendar years also differ accordingly. This phase is used to screen candidate model architectures, optimize model-specific hyperparameters, select features for feature-engineered models, and, for neural models, determine the training duration via validation-based early stopping. All model selection decisions are based exclusively on validation performance.

The final evaluation employs a walk-forward evaluation protocol \citep{hyndman2018forecasting}. Beginning with the initial training period defined during the Screening Phase, the training set is expanded by one year after each evaluation, and the model is retrained before generating a forecast for the subsequent year. This process is repeated over the five-year test period, producing five sequential out-of-sample forecasts for each country-crop dataset. Unlike a single hold-out split, the walk-forward protocol mimics the operational deployment of crop yield forecasting systems, where new historical observations become available each season while future information remains unavailable.

To ensure a fair and computationally tractable comparison, all hyperparameters selected during screening remain fixed throughout the walk-forward evaluation, thereby avoiding repeated tuning. 

Finally, to address stochastic variation during optimization, all trainable models are evaluated using five independent random seeds in the Walk-Forward Phase. Performance metrics are computed separately for each seed, and their average is reported as the final performance estimate. Deterministic models, such as the statistical baselines and the process-based crop model, are evaluated once.

\subsection{Hyperparameter optimization}
\label{sec:hyperparameter}

For each country--crop--model combination, we perform hyperparameter optimization during the Screening Phase using Tree-structured Parzen Estimator (TPE) Bayesian optimization implemented in Optuna \citep{akiba2019optuna}. Each optimization consists of 30 trials over three groups of model-specific hyperparameters: (i) \emph{input representation}, namely the number $k$ of mRMR-selected predictors for feature-based tabular models; (ii) \emph{model capacity and architecture}, including the embedding dimension and temporal encoder depth for neural models, tree depth and split criteria for tree-based models, and ensemble size for tabular foundation models; and (iii) \emph{optimization and regularization}, including learning rate, weight decay, batch size, dropout, and training duration for neural models, L2 regularization for ridge regression, and learning rate, subsampling ratio, and regularization parameters for gradient-boosting models.

For the feature-based models, mRMR uses the fixed F-test correlation difference (FCD) criterion \citep{zhao2019maximumrelevanceminimumredundancy}, in which feature relevance is quantified by the F-statistic and redundancy by pairwise correlations. Hyperparameter optimization selects only the number of retained predictors, $k$. During the Walk-Forward Phase, we re-fit mRMR independently at each forecast origin using only the corresponding training data while keeping the selected value of $k$ fixed. Neural models reuse the early-stopped training duration selected during screening. The statistical baselines and the process-based crop model do not require hyperparameter optimization.

\subsection{Country-specific modeling}
To account for differences in agro-ecological conditions and management practices, we train and tune each model independently for every country. Hyperparameter optimization is performed separately for each country, allowing architectural characteristics, such as the number of neural network layers, to adapt to the available data. Consequently, our evaluation compares modeling paradigms and model families (e.g., LSTMs, Transformers, and tree-based models) rather than fixed architectural configurations.

\subsection{Multi-dimensional evaluation}
\label{sec:multidim_eval}
Crop yield variability is largely driven by persistent spatial differences in productivity among administrative regions with distinct environmental and management conditions, whereas year-to-year variability within regions is comparatively smaller and primarily reflects interannual weather fluctuations and extreme events. Consequently, a model may achieve strong overall predictive performance by reproducing stable spatial yield patterns while failing to capture the temporal variations. To obtain a more comprehensive understanding of model behavior, this research evaluates predictive skill from four complementary perspectives: overall performance, spatial prediction, temporal prediction, and anomaly prediction.

\subsubsection{Overall prediction}
Following the existing crop yield research, we evaluate the overall performance using all region-year observations in the test set. This approach assesses our model's ability to predict absolute yields across both spatial and temporal dimensions using the coefficient of determination ($R^2$) and the normalized root mean squared error (NRMSE), both of which are widely used to evaluate model performance in this field \citep{archana2023survey, parashar2024integrated}.

\begin{table*}[!t]
\centering
\caption{Performance of representative models for each model family. Values are reported as median [IQR] across countries.}
\label{tab:family_representatives_crops}

\begin{tabular}{llccccc}
\toprule
&
&
\multicolumn{2}{c}{\textbf{Overall}} &
\textbf{Spatial} &
\textbf{Temporal} &
\textbf{Anomaly} \\
\cmidrule(lr){3-4}
\textbf{Family} &
\textbf{Model} &
\textbf{NRMSE $\downarrow$} &
\textbf{$\boldsymbol{R}^2$ $\uparrow$} &
\textbf{$\boldsymbol{r}$ $\uparrow$} &
\textbf{$\boldsymbol{r}$ $\uparrow$} &
\textbf{$\boldsymbol{r}$ $\uparrow$} \\
\midrule

\multicolumn{7}{l}{\textbf{Maize} (36 countries; 8{,}211 regions)}\\
\midrule

Baseline &
Trend &
25.4 [19.7, 33.7] &
0.18 [-0.33, 0.54] &
0.71 [0.58, 0.82] &
-0.27 [-0.53, 0.07] &
-0.10 [-0.17, 0.02] \\

Process-based &
LPJmL &
34.1$^\dagger$ [23.5, 50.5] &
-0.31$^\dagger$ [-1.05, -0.05] &
0.49$^\dagger$ [0.37, 0.65] &
0.22$^*$ [0.04, 0.49] &
0.08$^*$ [-0.01, 0.28] \\

Feature ML &
RF &
24.8 [17.9, 35.3] &
0.25 [-0.24, 0.52] &
0.72 [0.46, 0.82] &
0.27$^*$ [-0.09, 0.50] &
\textbf{0.16}$^*$ [-0.02, 0.25] \\

Foundation &
TabICL &
24.2$^*$ [17.0, 33.8] &
\textbf{0.25}$^*$ [-0.13, 0.52] &
\textbf{0.73} [0.56, 0.83] &
\textbf{0.33}$^*$ [0.07, 0.61] &
0.14$^*$ [0.04, 0.33] \\

Sequence &
Transformer &
\textbf{23.0} [16.3, 34.3] &
0.25 [-0.11, 0.50] &
0.70 [0.49, 0.80] &
0.14$^*$ [-0.04, 0.29] &
0.02$^*$ [-0.03, 0.19] \\

\midrule

\multicolumn{7}{l}{\textbf{Wheat} (27 countries; 3{,}878 regions)}\\
\midrule

Baseline &
Trend &
18.3 [15.4, 28.4] &
0.03 [-0.28, 0.55] &
0.73 [0.56, 0.89] &
-0.46 [-0.55, -0.23] &
-0.18 [-0.36, -0.03] \\

Process-based &
LPJmL &
23.5 [17.3, 33.8] &
0.05 [-0.61, 0.39] &
0.71 [0.53, 0.84] &
\textbf{0.34}$^*$ [0.01, 0.78] &
\textbf{0.18}$^*$ [0.05, 0.55] \\

Feature ML &
XGBoost &
16.7$^*$ [14.1, 25.4] &
0.31$^*$ [0.11, 0.61] &
\textbf{0.84} [0.61, 0.91] &
0.27$^*$ [0.03, 0.53] &
0.09$^*$ [0.01, 0.23] \\

Foundation &
TabDPT &
\textbf{15.9}$^*$ [14.2, 25.1] &
\textbf{0.46}$^*$ [0.08, 0.63] &
0.83$^*$ [0.68, 0.92] &
0.22$^*$ [-0.03, 0.66] &
0.13$^*$ [0.03, 0.47] \\

Sequence &
\textsc{TST} &
19.5 [15.6, 27.0] &
0.25 [-0.13, 0.44] &
0.77 [0.59, 0.90] &
-0.05$^*$ [-0.18, 0.30] &
-0.01$^*$ [-0.09, 0.16] \\

\bottomrule
\end{tabular}

\vspace{0.4em}

{\footnotesize
$^*$ significant improvement over Trend; $^\dagger$ significant degradation relative to Trend (country bootstrap, B = 10,000). Bold = best.
}

\end{table*}

\subsubsection{Spatial prediction:}
Spatial prediction evaluates whether a model reproduces the geographic distribution of yield across administrative regions within a given growing season. For each test year $t$, we compute the Pearson correlation between observed and predicted yields across regions:

\begin{equation}
r_{\mathrm{sp}}(t)=
r\!\left(
\{y_{r,t}\}_{r\in\mathcal{R}},
\{\hat y_{r,t}\}_{r\in\mathcal{R}}
\right),
\end{equation}

\noindent and report spatial prediction skill as the median of $r_{\mathrm{sp}}(t)$ across all test years.

\noindent This metric quantifies how well a model captures persistent differences in relative productivity between administrative regions, independent of absolute yield magnitude.

\subsubsection{Temporal prediction:}

Temporal prediction evaluates whether a model captures year-to-year yield variability within administrative regions. For each region $r$, we compute the Pearson correlation between observed and predicted yields across the available test years as follows:

\begin{equation}
r_{\mathrm{tm}}(r)=
r\!\left(
\{y_{r,t}\}_{t\in\mathcal{T}},
\{\hat y_{r,t}\}_{t\in\mathcal{T}}
\right),
\end{equation}

\noindent and report overall temporal prediction skill as the median of $r_{\mathrm{tm}}(r)$ across all regions.

\noindent This metric quantifies how well a model reproduces inter-annual yield fluctuations within regions.

\subsubsection{Anomaly prediction:}




To isolate deviations from long-term regional productivity, we center both observed and predicted yields using the observed regional mean,

\begin{equation}
\bar y_r=\frac{1}{|\mathcal{T}|}\sum_{t\in\mathcal{T}} y_{r,t},
\qquad
y'_{r,t}=y_{r,t}-\bar y_r,
\qquad
\hat y'_{r,t}=\hat y_{r,t}-\bar y_r.
\end{equation}

Anomaly skill is quantified as the Pearson correlation computed over all de-meaned region--year observations as follows:

\begin{equation}
r_{\mathrm{an}}=
r\!\left(
\{y'_{r,t}\}_{r,t},
\{\hat y'_{r,t}\}_{r,t}
\right).
\end{equation}

\noindent This metric evaluates whether the model correctly predicts years with above- or below-average yields, independently of persistent regional productivity differences.

Overall, we evaluate the spatial, temporal, and anomaly prediction using Pearson's correlation coefficient, as these analyses focus on agreement in spatial and temporal patterns rather than absolute yield magnitudes.



\section{Results}

This section provides only a summary of the results from all trained models. The complete benchmark—including all evaluated models, crops, prediction horizons, performance metrics, qualitative analyses, and explainability visualizations—is available through the accompanying interactive dashboard at \href{https://wur-ai.github.io/AgML-CY-Bench-dashboard}{\textcolor{blue}{https://wur-ai.github.io/AgML-CY-Bench-dashboard}}. 

\subsection{Comparison of modeling paradigms}
We first compare representative models from each modeling paradigm. Representatives are selected based on the lowest median NRMSE across countries. Table~\ref{tab:family_representatives_crops} summarizes their performance across the four evaluation dimensions introduced in Section~\ref{sec:multidim_eval}. %

Several observations emerge. First, overall predictive performance differs between maize and wheat. For wheat, tabular approaches consistently achieve the strongest overall performance, outperforming the statistical, process-based, and sequence baselines on both NRMSE and $R^2$. For maize, differences between the strongest data-driven paradigms are comparatively modest: the sequence representative achieves the lowest median NRMSE, while the strongest tabular model achieves the highest overall $R^2$. Overall, feature-engineered machine learning and tabular foundation models perform similarly, particularly for maize, while deep sequence models do not consistently outperform either approach.



Second, all model families achieve substantially higher spatial than temporal skill. Finally, anomaly prediction remains challenging across all modeling paradigms, indicating that forecasting departures from a region’s expected productivity continues to be one of the most difficult aspects of sub-national crop yield prediction.


The Trend baseline illustrates why evaluating multiple aspects of predictive skill matters. Despite achieving competitive overall accuracy and strong spatial correlation, Trend exhibits negative temporal and anomaly correlations for both crops. This shows that models can achieve good pooled prediction error largely by reproducing long-term regional productivity patterns, without correctly identifying favorable and unfavorable growing seasons. The strongest tabular models substantially improve temporal and anomaly skill relative to Trend, although anomaly prediction remains considerably more difficult than spatial prediction.

Finally, the process-based LPJmL model exhibits a distinct performance profile. Although its overall predictive accuracy is generally lower than that of the strongest data-driven approaches, it achieves comparatively strong temporal skill, with the highest median temporal (and anomaly) correlation for wheat. This suggests that process-based models capture aspects of environmentally driven inter-annual variability despite their comparatively lower overall predictive accuracy.

\subsection{Country-level benefit of data-driven models}
We assess the consistency of model performance across countries by comparing, for each country, the best-performing data-driven model (feature-engineered, sequence, or foundation) with the best traditional baseline (Trend, Average, or LPJmL). The relative improvement is defined as:

\begin{equation}
\Delta_{\mathrm{AI}}
=
\frac{
\min\limits_{m \in \mathcal{M}_{\mathrm{trad}}}
\mathrm{NRMSE}(m)
-
\min\limits_{m \in \mathcal{M}_{\mathrm{AI}}}
\mathrm{NRMSE}(m)
}{
\min\limits_{m \in \mathcal{M}_{\mathrm{trad}}}
\mathrm{NRMSE}(m)
},
\label{eq:ai_gain}
\end{equation}

\begin{figure*}[!t]
    \centering
    \begin{minipage}[t]{0.48\textwidth}
        \centering
        \textbf{(a) Maize}\\[0.4em]
        \includegraphics[width=\linewidth]{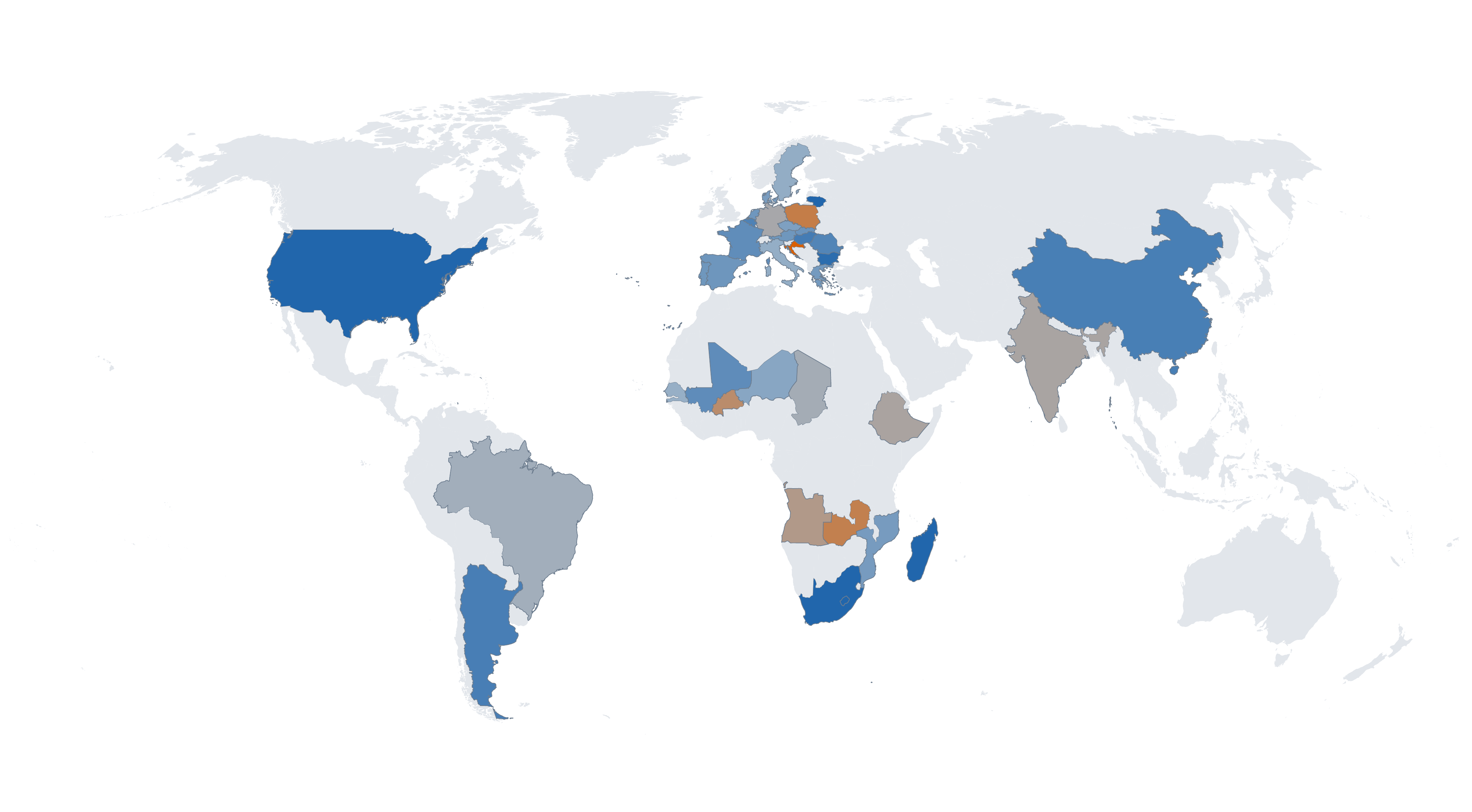}
    \end{minipage}\hfill
    \begin{minipage}[t]{0.48\textwidth}
        \centering
        \textbf{(b) Wheat}\\[0.4em]
        \includegraphics[width=\linewidth]{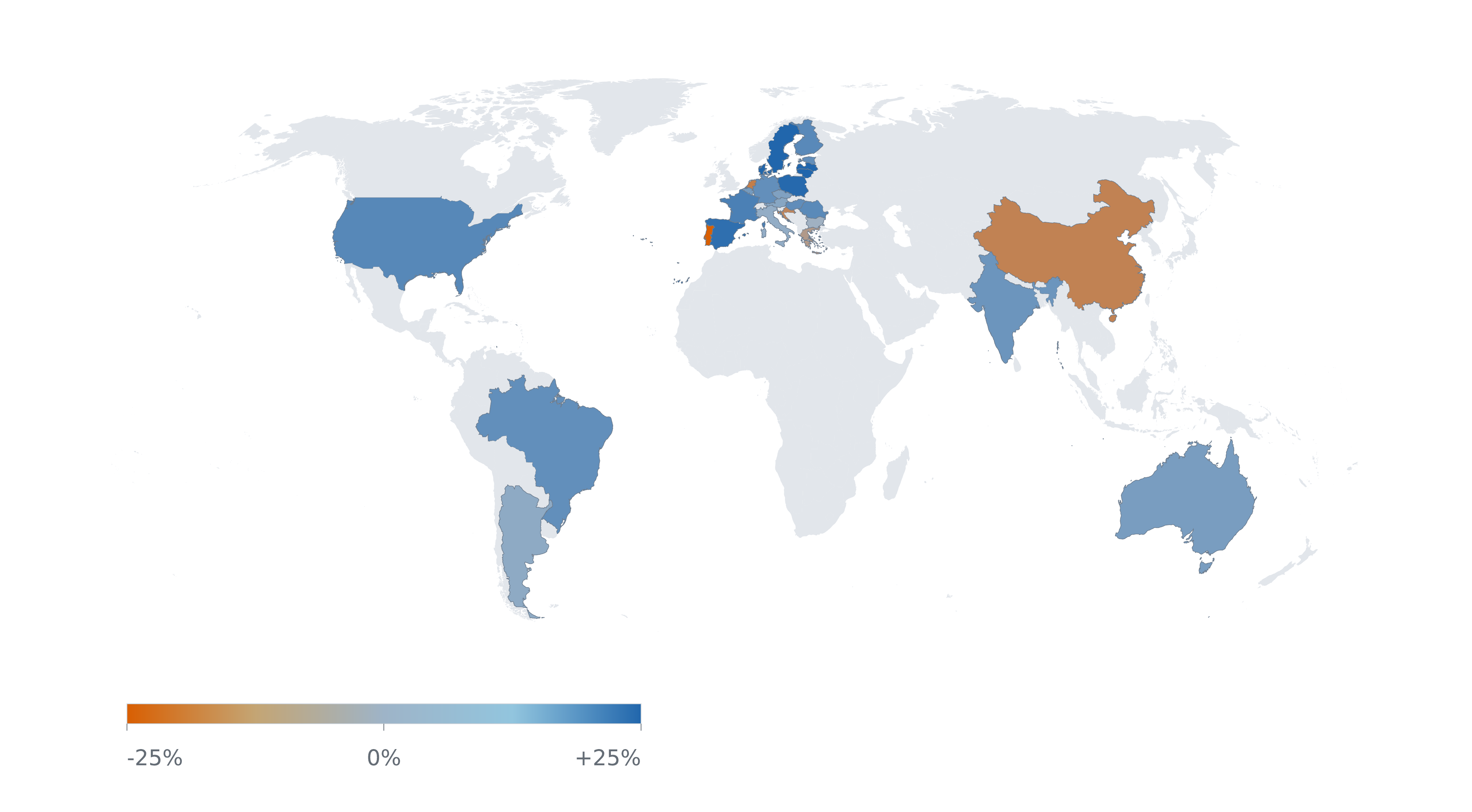}
    \end{minipage}
    \caption{Relative reduction in NRMSE of the best data-driven paradigm compared with the best traditional approach. Positive values indicate improved predictive performance through data-driven modeling.}
    \Description{Relative reduction in NRMSE of the best data-driven paradigm (feature-engineered, foundation, or sequence) compared with the best traditional approach (statistical or process-based). Positive values indicate improved predictive performance through data-driven modeling. Results are shown separately for (a) maize and (b) wheat.}
    \label{fig:ai_gain_maps}
\end{figure*}

\noindent Figure~\ref{fig:ai_gain_maps} summarizes the country-level benefit of data-driven models. For wheat, data-driven models outperform the strongest traditional baseline in 21 of 27 countries (78\% (95\% CI: 63–93\%)), yielding a median NRMSE reduction of  11.9\% (95\% CI: 4.6–13.8\%). For maize, improvements are slightly smaller and more heterogeneous: data-driven models outperform the traditional baselines in 26 of 36 countries (72\% (95\% CI: 58–86\%)), with a median NRMSE reduction of 8.8\% (95\% CI: 2.3–12.4\%). 

While the primary benchmark evaluates end-of-season forecasting to enable comparison across all modeling paradigms, Appendix \ref{appendix:mid_season} additionally reports representative-model performance for mid-season forecasting. Although predictive performance generally improves toward the end of the growing season, the supplementary analysis demonstrates that useful crop yield forecasts are already possible by mid-season.

\section{Discussion and conclusion}
This benchmark provides several insights into the current state of AI for sub-national crop yield forecasting that extend beyond differences in predictive accuracy between individual models.

First, data-driven approaches generally achieve the strongest predictive performance. Conventional machine learning models trained on engineered agronomic features remain highly competitive with tabular foundation models, which use the same feature representation but benefit from large-scale pretraining. Their similar performance suggests that the engineered representation already captures much of the predictive information available in current large-scale crop yield datasets, leaving little room for improvement through pretraining. In contrast, sequence models, which learn directly from raw daily observations, do not consistently outperform either approach. This suggests that, under the relatively small data regimes available for many country--crop combinations, engineered agronomic features provide a stronger inductive bias than learning temporal representations directly from raw observations.

Second, the benchmark reveals a clear distinction between predicting spatial and temporal variability. Across all modeling paradigms, spatial prediction skill is substantially higher than temporal prediction skill, indicating that reproducing persistent regional productivity differences is considerably easier than predicting inter-annual variability within individual regions. This suggests that current models successfully capture stable differences associated with long-term environmental and agronomic conditions, whereas transient environmental drivers, including weather extremes, and water stress, remain considerably more difficult to model. Since the impact of extreme events depends strongly on their timing relative to crop phenology, further progress will likely require richer observations of rare extreme events and their interaction with crop development. This is particularly important for agricultural early-warning systems, as anticipating year-to-year anomalies is more valuable than reproducing long-term gradients. 

The contrasting behavior of the Trend and LPJmL baselines further illustrates this distinction. Trend achieves competitive pooled accuracy despite strongly negative temporal skill, demonstrating that good overall performance can largely be obtained by reproducing persistent productivity gradients without explaining inter-annual variability. Conversely, LPJmL sacrifices overall predictive accuracy while exhibiting comparatively strong temporal skill, suggesting that process-based models capture environmentally driven year-to-year responses despite limitations arising from coarse spatial resolution, global parameterization, and limited biophysical process representation especially for extreme event responses. 

Together, these findings demonstrate that pooled accuracy alone provides an incomplete assessment of crop yield forecasting performance, and spatial, temporal and anomaly detection skill needs to be evaluated separately. When model evaluation focuses only on aggregate error, the complementary skills of individual models to estimate crop responses is largely obscured.

The complementary strengths of the data-driven, statistical, and process-based approaches further suggest that hybrid modeling frameworks may provide a promising avenue to improve regional crop yield prediction by combining physically consistent process representations with data-driven learning. Learning process-based residuals or designing physics-informed neural networks constrained with crop physics offers the potential to leverage mechanistic crop dynamics while accounting for observational constraints and non-linear relationships \citep{Maestrini2022, oikonomidis2022hybrid, kallenberg2023integrating, Han2023, han2025, Jorges2026Regional}.

Third, anomaly prediction remains challenging across all modeling paradigms, indicating that current forecasting systems only partially capture the processes driving inter-annual yield variability. Beyond model comparison, these findings also provide insight into the nature of regional crop yield variability itself: persistent spatial productivity patterns can largely be recovered from currently available predictors, whereas inter-annual fluctuations remain only partially explained. One plausible explanation is that current large-scale benchmark datasets contain only limited information on management practices, irrigation, cultivar selection, pests, diseases, and local extreme events, all of which may contribute to this unexplained variability. This highlights a limitation of current forecasting systems, which rely primarily on weather and vegetation indicators while only implicitly representing many of the processes influencing crop production. Taken together,  future progress in large-scale crop yield forecasting will require improved characterization of the environmental and management processes driving inter-annual variability, including Earth observation to better characterize crop phenology and stress dynamics, improved agricultural management information, and advances in modeling that better capture the complex relationships between these factors.

The supplementary analysis indicates that useful yield forecasts are already possible by mid-season. This suggests that much of the information predictive of final yield is already available by mid-season, while the incremental value of late-season observations may be reduced by senescence and inter-annual phenological variability.

Beyond these empirical findings, this work demonstrates the value of standardized benchmarking for agricultural AI. By extending CY-Bench with an additional quality-control layer, a walk-forward evaluation protocol, and a harmonized implementation of diverse modeling paradigms, we enable a more rigorous comparison across fundamentally different forecasting approaches. In particular, driving both LPJmL and the data-driven models with identical meteorological forcing, soil properties, and crop distributions minimizes differences arising from input data and evaluation procedures, allowing more direct comparison of model behavior. Such standardized benchmark frameworks are essential for transparent evaluation of future crop yield forecasting methods.

Finally, the benchmark highlights several priorities for future research. Strong feature-engineered baselines should remain part of future evaluations, as they leverage domain knowledge and continue to rival considerably more complex sequence and foundation models. Greater emphasis should also be placed on improving temporal and anomaly prediction rather than further optimizing already strong spatial prediction. More broadly, continued progress is likely to depend not only on improved learning algorithms, but also on richer observations of crop management and transient environmental processes that better explain inter-annual yield variability. Unlike process-based crop models, which explicitly simulate phenological development and evaluate environmental stresses relative to crop growth stages, data-driven models mostly rely on calendar time and must infer these relationships from data. 
Therefore, future developments may consider learning representations that capture time-series relative to crop phenological stages and thus better capture stage-specific responses to weather and water stress, offering a promising direction for future crop forecasting models and hybrid approaches.


\section{Limitations and ethical considerations}
Several limitations should be noted. First, although CY-Bench covers a large number of countries and two major crops, it does not include management variables such as fertilization, irrigation, cultivar choice, tillage, or pest pressure, all of which influence yield variability. Such data are not readily available at continental scales and across countries, but are typically collected only in local-scale experimental agricultural studies. Second, the process-based benchmark is represented by a single globally parameterized LPJmL configuration and should therefore not be interpreted as representative of all process-based crop models. Finally, although the benchmark spans a diverse collection of modeling paradigms, new foundation and multimodal models continue to emerge and should be incorporated in future benchmark releases.

Crop yield forecasts can inform agricultural planning, food security assessments, and commodity markets. They should therefore be used as decision-support tools rather than as standalone decision-making systems, and their predictions should be interpreted together with agronomic expertise and local knowledge. We hope that our benchmark contributes to more transparent and reproducible evaluation of crop yield forecasting models while encouraging future work on improved data quality, richer environmental and management predictors, and robust evaluation protocols.

\section*{Generative AI usage}
Generative AI tools were used to support software development and manuscript preparation. Cursor AI assisted with implementation of the dashboard. AI-generated content was reviewed and validated by the authors, who remain responsible for the study design, methodology, results, and conclusions.

\section*{Data and code availability}
The experiments are conducted on the publicly available CY-Bench dataset \citep{kallenberg2026cybench}. We provide an interactive dashboard with results across datasets, model families, prediction horizons, and evaluation metrics at \href{https://wur-ai.github.io/AgML-CY-Bench-dashboard}{\textcolor{blue}{https://wur-ai.github.io/AgML-CY-Bench-dashboard}}.

The source code for this work is available at \href{https://github.com/WUR-AI/AgML-CY-Bench-Modeling}{\textcolor{blue}{https://github.com/WUR-AI/AgML-CY-Bench-Modeling}}.

\begin{acks}
We thank \href{https://www.agml.org}{AgML}, the Machine Learning team of the  \href{http://www.agmip.org}{Agricultural Model Intercomparison and Improvement Project}. This work was supported by the EU Horizon projects AI Foundation Models in Agricultural Sciences (101293777) and Ambrosia (101181300), and the Digital Europe Programme through AgrifoodTEF (101100622).
\end{acks}

\bibliographystyle{ACM-Reference-Format}
\bibliography{cybench}

\appendix
\section{Mid-season forecasting}
\label{appendix:mid_season}
The primary benchmark evaluates end-of-season forecasting, allowing comparison across all modeling paradigms. To assess the practical feasibility of operational forecasting before harvest, we additionally evaluate representative models using only predictor information available halfway through the growing season. Table \ref{tab:mid_season} summarizes the change in predictive performance relative to end-of-season forecasting across the four evaluation dimensions.

\begin{table*}[!t]
\centering
\caption{Difference between mid-season ($\sim$50\% of the growing season observed) and end-of-season ($\sim$100\%) performance for the representative model of each family. Values denote the median $\Delta$ [IQR] across countries. Positive $\Delta$ indicates improved end-of-season performance.}
\label{tab:mid_season}

\begin{tabular}{llccccc}
\toprule
&
&
\multicolumn{2}{c}{\textbf{Overall}} &
\textbf{Spatial} &
\textbf{Temporal} &
\textbf{Anomaly} \\
\cmidrule(lr){3-4}
\textbf{Family} &
\textbf{Model} &
\textbf{$\Delta$NRMSE} &
\textbf{$\Delta R^2$} &
\textbf{$\Delta r$} &
\textbf{$\Delta r$} &
\textbf{$\Delta r$} \\
\midrule

\multicolumn{7}{l}{\textbf{Maize} (36 countries; 8{,}211 regions)}\\
\midrule

Baseline &
Trend &
0.0 [0.0, 0.0] &
0.00 [0.00, 0.00] &
0.00 [0.00, 0.00] &
0.00 [0.00, 0.00] &
0.00 [0.00, 0.00] \\

Process-based &
LPJmL &
N/A &
N/A &
N/A &
N/A &
N/A \\

Feature ML &
RF &
0.3 [-0.7, 1.3] &
0.01 [-0.03, 0.09] &
0.00 [-0.04, 0.02] &
0.06 [-0.10, 0.16] &
0.01 [-0.05, 0.07] \\

Foundation &
TabICL &
0.4$^*$ [-1.4, 2.4] &
0.03$^*$ [-0.07, 0.10] &
0.03 [0.00, 0.07] &
0.12$^*$ [-0.04, 0.26] &
0.03 [-0.03, 0.15] \\

Sequence &
Transformer &
-0.1 [-1.2, 1.8] &
0.00 [-0.06, 0.14] &
0.01 [-0.03, 0.02] &
0.01 [-0.05, 0.13] &
0.02 [-0.04, 0.07] \\

\midrule

\multicolumn{7}{l}{\textbf{Wheat} (27 countries; 3{,}878 regions)}\\
\midrule

Baseline &
Trend &
0.0 [0.0, 0.0] &
0.00 [0.00, 0.00] &
0.00 [0.00, 0.00] &
0.00 [0.00, 0.00] &
0.00 [0.00, 0.00] \\

Process-based &
LPJmL &
N/A &
N/A &
N/A &
N/A &
N/A \\

Feature ML &
XGBoost &
0.1 [-0.4, 1.6] &
0.00 [-0.02, 0.09] &
0.00 [-0.01, 0.02] &
0.09 [-0.08, 0.16] &
0.03 [-0.05, 0.12] \\

Foundation &
TabDPT &
1.5 [-0.4, 3.2] &
0.05 [-0.02, 0.22] &
0.02$^*$ [-0.01, 0.05] &
0.13$^*$ [-0.07, 0.33] &
0.04$^*$ [-0.02, 0.24] \\

Sequence &
TST &
1.1$^*$ [-0.5, 2.5] &
0.08$^*$ [-0.03, 0.26] &
0.01$^*$ [0.01, 0.05] &
0.11$^*$ [-0.08, 0.23] &
0.02 [-0.09, 0.13] \\

\bottomrule
\end{tabular}

\vspace{0.4em}

{\footnotesize
$^*$ end-of-season significantly better than mid-season (country bootstrap, $B=10{,}000$). Positive $\Delta$ indicates improved end-of-season performance. LPJmL is omitted because no mid-season forecasts are available.
}

\end{table*}

Across both crops, predictive performance generally improves when observations from the complete growing season are available. Nevertheless, the relatively small performance differences indicate that substantial predictive skill is already available at mid-season, suggesting that useful yield forecasts can be produced several months before harvest.




\end{document}